\documentclass{article}
\usepackage{iclr2027_conference,times}

\usepackage{amsmath,amsfonts,bm}

\def\eqref#1{equation~\ref{#1}}

\def\1{\bm{1}}

\DeclareMathAlphabet{\mathsfit}{\encodingdefault}{\sfdefault}{m}{sl}
\SetMathAlphabet{\mathsfit}{bold}{\encodingdefault}{\sfdefault}{bx}{n}

\usepackage{hyperref}
\usepackage{url}
\usepackage{booktabs,multirow}
\usepackage{amsmath,amssymb}
\usepackage[protrusion=false]{microtype}
\usepackage{graphicx}
\usepackage{algorithm,algpseudocode}
\usepackage{placeins}
\usepackage{array}
\usepackage{xcolor,colortbl,tabularx}

\usepackage[most]{tcolorbox}
\definecolor{orsdPromptBlue}{HTML}{244A73}
\definecolor{orsdPromptGreen}{HTML}{277061}
\definecolor{orsdPromptPurple}{HTML}{605384}
\newcommand{\promptvar}[1]{\textcolor{orsdPromptBlue}{\texttt{\{#1\}}}}
\newtcolorbox{orsdprompt}[2]{enhanced,breakable,
 title after break={#2 (continued)},
 colback=#1!3!white,colframe=#1,colbacktitle=#1,coltitle=white,
 fonttitle=\bfseries,title={#2},boxrule=0.6pt,arc=2mm,
 left=9pt,right=9pt,top=7pt,bottom=7pt,
 before skip=10pt,after skip=12pt,fontupper=\normalsize,
 before upper={\raggedright\setlength{\parskip}{5pt}}}

\newcommand{\method}{\textsc{Online Rubric Self-Distillation}}
\newsavebox{\orsdMainResultsBox}
\newlength{\orsdModelColumnOffset}

\newcommand{\sg}{\operatorname{sg}}

\title{Using Experience Twice:
Dynamic Process Rubrics for Privileged Token-Level Credit Assignment}
\author{%
Shumao Wu\textsuperscript{$1,3$}, Kongcheng Zhang\textsuperscript{$2$}, Jianqiang Huang\textsuperscript{$1,3$}, Baisheng Lai\textsuperscript{$1,3$}\thanks{Corresponding author: \texttt{bslai@cnic.cn}} \\
\textsuperscript{$1$}University of Chinese Academy of Sciences \\
\textsuperscript{$2$}Zhejiang University \\
\textsuperscript{$3$}Computer Network Information Center, Chinese Academy of Sciences\\
\texttt{smwu@cnic.cn},
\texttt{zhangkc@zju.edu.cn},
\texttt{jqhuang@cnic.cn},
\texttt{bslai@cnic.cn}
}

\iclrfinalcopy
\begin{document}
\maketitle
\lhead{}

\begin{abstract}
Agent trajectories often contain both successful decisions and failed attempts, yet optimization with sparse trajectory-level rewards struggles to identify which critical decisions should be reinforced or suppressed. We observe that meaningful distinctions among on-policy trajectories can be distilled into natural-language process rubrics and used as process supervision signals. Based on this insight, we propose \method{} (ORSD), which leverages self-derived process rubrics for dense token-level supervision, thereby augmenting outcome-centric reinforcement learning. Specifically, the current policy independently generates multiple rubric bundles at each step according to the same group of trajectories, extracting candidate process rubrics from cross-trajectory behavioral differences. A frozen verifier then evaluates all rubric items against each trajectory in the group, and selects one candidate bundle based on bundle-level alignment between rubric satisfaction and task success. ORSD treats the selected rubric bundle as privileged information during training, then compares the log-likelihoods of the original tokens with and without the rubric. The resulting likelihood gap provides token-specific process credit, which is added to the original trajectory-level GRPO advantage through additive advantage shaping. ORSD thus bridges trajectory-level outcome supervision and fine-grained process supervision, enabling token-level credit assignment using process signals. Extensive experiments on ALFWorld and WebShop show that ORSD accelerates policy improvement and boosts final task performance for both Qwen3-1.7B and Qwen3-4B under the same environment-interaction budget. Our code is available at \url{https://github.com/AnnaXChun/ORSD}.

\end{abstract}
\section{Introduction}
\label{sec:introduction}

Large language model (LLM) agents have shown strong capabilities in web interaction~\citep{zhou2024webarena} and software engineering~\citep{yang2024sweagent}, with applications extending to embodied exploration~\citep{wang2024voyager}. Yet reliable decision-making in interactive environments remains challenging: agents must plan sequences of actions, explore unfamiliar situations, and reflect on feedback to revise unsuccessful strategies. Recent progress in reinforcement learning with verifiable rewards (RLVR) offers a promising training approach. DeepSeek-R1~\citep{guo2025deepseek} shows that verifiable reward feedback can substantially improve reasoning performance, with reflection and self-verification emerging during reinforcement learning. Building on these advances, agentic RL has also adopted Group Relative Policy Optimization (GRPO)~\citep{shao2024deepseekmath} to train agents through environment interaction, as demonstrated by RAGEN~\citep{wang2025ragen} in multi-turn tasks.

Despite these advances, relying solely on outcome rewards provides limited guidance about which intermediate decisions deserve credit. Under trajectory-level GRPO, all policy-generated tokens within a trajectory share the same advantage, so useful actions, redundant steps, and recoverable mistakes receive the same learning signal. To assign credit to individual steps, GiGPO~\citep{feng2025gigpo} groups actions from matching environment states across trajectories and estimates step-level advantages through within-group comparisons. This refines credit across decisions, although tokens within each action still share the same step-level signal. More recently, natural-language rubrics have emerged as another source of supervision by making task-specific evaluation criteria explicit. Rubrics as Rewards~\citep{gunjal2025rubrics} use these criteria to construct reward signals. RLCF~\citep{viswanathan2025checklists} constructs instruction-specific checklists in advance and aggregates criterion-level judgments into a response score, while OpenRubrics~\citep{liu2026openrubrics} derives rubrics from existing preference pairs and trains a rubric-guided reward model for preference optimization. Although rubric-based methods introduce richer evaluation criteria, their feedback usually still remains at the trajectory level, leaving credit assignment to specially individual tokens unresolved. Moreover, their rubric construction is separate from the policy's ongoing learning process, so the criteria are not continually updated by the updated policy.

These limitations motivate the use of rubrics as privileged context for finer-grained credit assignment. Our core insight is that comparing the log-probabilities assigned to the same sampled tokens with and without rubrics can turn rubric-based guidance into token-level advantages. Such signals can refine trajectory-level advantages, allowing tokens within the same trajectory to receive different credits. Moreover, as the policy evolves during training, its exploration patterns and failure modes may also change. Generating rubrics online from on-policy trajectories allows them to reflect the policy's current behavior, capturing task-specific constraints, failure patterns, and successful action sequences that static rubrics may overlook. This enables the credit assignment signal to adapt to the evolving policy rather than relying on fixed, predefined criteria. However, jointly deriving rubrics from on-policy experience and leveraging them as privileged context for token-level credit assignment remains underexplored.

In this work, we introduce \method{} (ORSD), a framework that turns on-policy experience into adaptive process supervision for token-level credit assignment. ORSD reuses these trajectories to construct process supervision in three steps. First, conditioned on both task prompt and the group of trajectories, the same policy independently generates multiple rubric bundles, each containing multiple criteria for evaluating intermediate states along the trajectories. Second, ORSD employs a verifier to assess the satisfaction of each rubric criterion on every trajectory, producing a satisfaction profile for each rubric bundle. It then measures the alignment between each satisfaction profile and the outcome rewards, and selects the rubric bundle with the highest correlation as privileged information. Third, ORSD augments the prompt with the selected privileged information and recomputes the token likelihoods of the sampled trajectories. The gap between the original and augmented token likelihoods is then used to compute the token-level advantage. As a result, ORSD transforms a group of on-policy trajectories into token-level credit, requiring neither additional rollouts nor additional environment interactions.

Our contributions are summarized as follows:
\begin{itemize}
    \item{We propose \method{} (ORSD), a framework that uses online-generated process rubrics to derive token-level credit. These rubrics are distilled from the policy’s own rollouts and serve as privileged information that augments the replay context, with token-level log-probability differences between replay with and without the rubrics underpinning this fine-grained credit allocation.}
    \item{To address the quality variation among generated rubrics, we propose a rubric-reward alignment mechanism that filters out low-value rubrics by measuring the alignment between rubric satisfaction and task success, thereby enhancing the quality of token-level credit.}
    \item{Experiments on ALFWorld and WebShop with Qwen3-1.7B and Qwen3-4B show that ORSD improves overall success over GRPO under the same environment-interaction budget. Ablations support the contributions of online rubric generation and rubric–reward alignment.}
\end{itemize}

\section{Related Work}
\label{sec:related}

\paragraph{\textbf{RLVR for LLM Agents.}}
Reinforcement learning with verifiable rewards (RLVR) uses automatically verified outcomes to train language models. PPO and GRPO are commonly used for policy optimization, with GRPO estimating relative advantages within groups of sampled responses~\citep{schulman2017ppo,shao2024deepseekmath}. For LLM agents, training involves multiple rounds of interaction, during which policy's decisions affect subsequent observations and the final reward. RAGEN examines learning dynamics in multi-turn agent RL, and Agent Lightning provides a framework for training agents from their execution traces~\citep{wang2025ragen,luo2025agentlightning}. The environment interaction also supports additional learning signals. EnvRL uses state transitions for representation learning, while T$^2$PO adjusts exploration through uncertainty estimates and turn-level resampling~\citep{wang2026envrl,wang2026t2po}.

\paragraph{Credit Assignment in Agentic RL.}
Credit assignment assigns different credit to individual steps or tokens in an agent trajectory. GiGPO groups steps across trajectories by state similarity and estimates relative advantages within each group~\citep{feng2025gigpo}. HGPO further groups steps according to their interaction contexts~\citep{he2026hgpo}. ArCHer uses rewards from subsequent interactions to estimate the value of the entire response generated at the current turn, then uses this estimate to update the policy that generates the response~\citep{zhou2024archer}. POAD further considers the role of individual tokens within an action: it treats action generation as a sequence of token-level decisions and estimates future rewards for each token, providing a finer-grained training signal~\citep{wen2024poad}. SC-GRPO conditions on verified solutions and weights token gradients according to the resulting distributional divergence~\citep{shan2026scgrpo}.

\paragraph{Natural-language Rubrics.}
A natural-language rubric is a set of written criteria specifying which aspects of a response or reasoning process to evaluate. Prior work uses rubrics to provide feedback at different levels: Rubrics as Rewards assigns a reward to the entire response based on rubric satisfaction~\citep{gunjal2025rubrics}; Rubrics to Tokens estimates the relevance of each criterion to individual response tokens, providing token-level feedback~\citep{xu2026rubricstotokens}; and Step-wise Rubric Rewards checks whether each reasoning step satisfies the criteria and assigns step-level rewards~\citep{xie2026stepwise}. These methods use rubrics to construct fine-grained supervision, with criteria generated offline and held fixed. We further investigate how to generate rubrics online that better match the current policy and use them for token-level credit assignment.

\paragraph{On-policy Distillation with Privileged Context.}
Knowledge distillation trains a student model using teacher predictions, with supervision applied to individual tokens or complete sequences~\citep{hinton2015distilling,kim2016sequence}. Training can also use data generated by the learner itself: DAgger collects supervision at states visited by the learner's policy~\citep{ross2011dagger}, while GKD~\citep{agarwal2023gkd} trains on student-generated sequences with supervision provided by a teacher. OPSD~\citep{zhao2026opsd} uses the same model for self-distillation: the model first generates a response, then computes its log-probability given additional information to provide a training signal for the original policy. This additional information is called privileged context. Prior work also extracts such information from task execution records: Skill-SD~\citep{wang2026skillsd} summarizes experience from completed trajectories into natural-language skills, while SDPO~\citep{hubotter2026sdpo} uses environmental feedback and successful attempts for self-distillation.

\section{Method}
\label{sec:method}

\begin{figure}[!htbp]
\centering
\includegraphics[width=\textwidth]{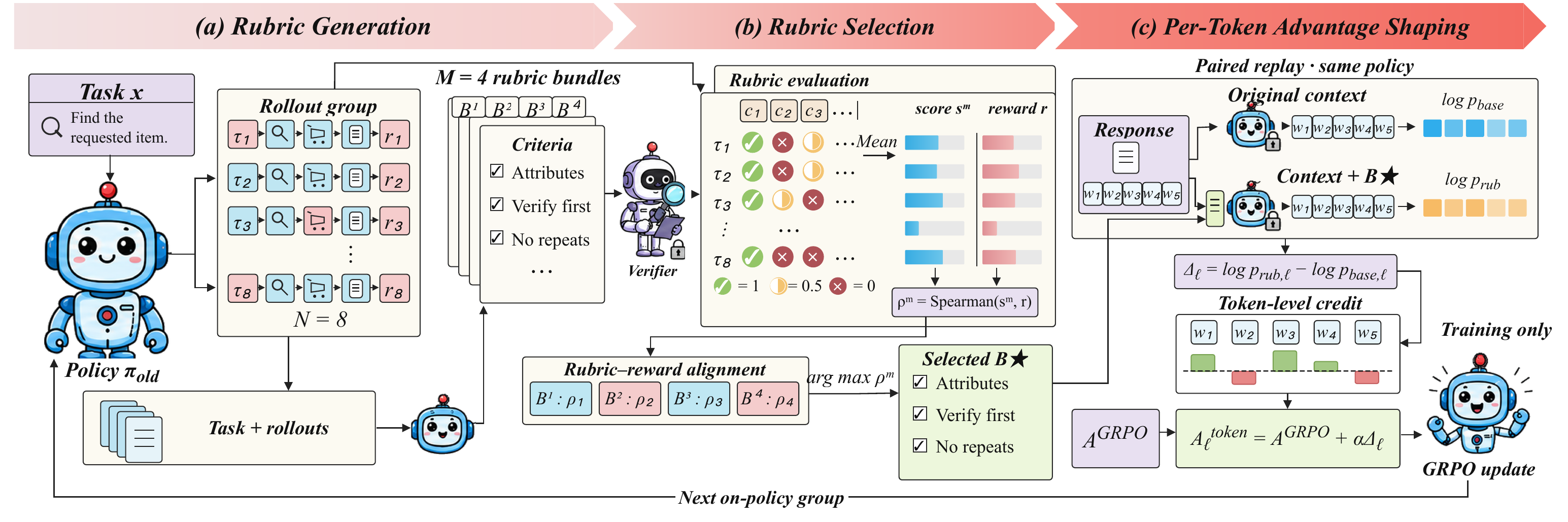}
\caption{The overview of ORSD. (a) For each task, the policy first samples a group of trajectories. It then takes the task description and these trajectories as inputs to generate multiple rubric bundles. (b) Each Rubric bundle is evaluated on every trajectory, the rubric bundle with the largest rubric-reward alignment score is selected. (c) The selected rubric bundle is used to augment the context and compute the per-token likelihood gap between augmented and non-augmented contexts, enabling per-token advantage shaping of the original trajectory-level advantage.}
\label{fig:overview}
\end{figure}

In this section, we present the proposed \method{} (ORSD), a framework that turns on-policy rollouts into rubrics and then into token-level credit signals. As shown in Figure~\ref{fig:overview}, at each training step, the current policy first samples a group of trajectories for the same task and obtains a reward for each trajectory. It then generates candidate rubric bundles from the task description and these trajectories. The rubric-reward alignment mechanism selects the bundle whose rubric satisfaction scores correlate most strongly with the rewards across the group. Finally, the token-wise log-likelihood difference between rollouts with and without the selected bundle enables per-token advantage shaping of the original trajectory-level advantage. In this way, ORSD reuses the same on-policy trajectories for rubric generation and privileged paired replay to derive token-specific credit without introducing additional environment interactions.

\subsection{Preliminaries}
Let $\pi_{\mathrm{old}}$ denote the policy used to sample trajectories in the current step. For a task $x$, ORSD samples $N$ trajectories:
\begin{equation}
\mathcal T=\{\tau_i\}_{i=1}^{N},
\qquad \tau_i\sim\pi_{\mathrm{old}}(\cdot\mid x).
\label{eq:trajectory-group}
\end{equation}
The reward for trajectory $\tau_i$ is $r_i=R_{\mathrm{train}}(\tau_i)$, as defined in Section~\ref{sec:reward-design}. As in GRPO, ORSD normalizes rewards within each task group to obtain a trajectory-level advantage, which serves as the base advantage for subsequent token-level advantage shaping~\citep{shao2024deepseekmath}:
\begin{equation}
A_i^{\mathrm{GRPO}}
=\frac{r_i-\mu_r}{\sigma_r+\varepsilon_{\mathrm{num}}}.
\label{eq:grpo-adv}
\end{equation}
Here, $\mu_r$ and $\sigma_r$ are the mean and sample standard deviation of the group rewards. ORSD will add a token-level correction to this advantage.

\subsection{Online Rubric Bundle Generation and Selection}
\label{sec:rubric-generation}
Building on the same-task rollout group mentioned above, ORSD further exploits the behavioral distinctions across trajectories to construct process supervision. For example, in WebShop, one trajectory may verify product attributes before purchasing, whereas another may skip this step and end up selecting an item that does not satisfy the user's request. These behavioral differences can be summarized as natural-language process rubrics. At each training step, we provide the current policy with the task description $x$, the sampled trajectory group $\mathcal T$, and a rubric-generation instruction $p_r$, and independently sample $M$ candidate rubric bundles:
\begin{equation}
\mathcal B^{(m)}\sim
\pi_{\mathrm{old}}(\cdot\mid x,\mathcal T,p_r),
\qquad m=1,\ldots,M.
\label{eq:rubric-generation}
\end{equation}
Each bundle contains a set of process criteria, denoted by $\mathcal B^{(m)}=\{\kappa_k^{(m)}\}_{k=1}^{K^{(m)}}$, where the superscript $(m)$ indexes the candidate bundle, $k$ indexes a criterion within it, and $K^{(m)}$ is its number of criteria.

Different from prior works, our method produces candidate rubric bundles online using the policy instead of human-defined criteria or generation by stronger models. This mechanism enables the generated rubrics to capture the dynamics of on-policy trajectories, while the challenge is that candidate rubric bundles exhibit variable quality. To address this, we select one rubric bundle among the candidates based on a rubric–reward alignment score. Specifically, a verifier $V$ first scores each trajectory against each criterion $\kappa_k^{(m)}$:
\begin{equation}
S_{i,k}^{(m)}=V(x,\tau_i,\kappa_k^{(m)})\in[0,1].
\label{eq:satisfaction}
\end{equation}
For trajectory $i$, we average the criterion-level scores $S_{i,k}^{(m)}$ to obtain the bundle-level satisfaction score $s_i^{(m)}$:
\begin{equation}
s_i^{(m)}=\frac{1}{K^{(m)}}\sum_{k=1}^{K^{(m)}}S_{i,k}^{(m)}.
\label{eq:bundle-satisfaction}
\end{equation}

We then compute the Spearman correlation between bundle satisfaction and outcome rewards across all $N$ trajectories for the task:
\begin{equation}
\rho^{(m)}=\operatorname{Spearman}\!\left(
\{s_i^{(m)}\}_{i=1}^{N},\{r_i\}_{i=1}^{N}
\right).
\label{eq:bundle-score}
\end{equation}
This correlation measures the agreement between the rankings in $\{s_i^{(m)}\}_{i=1}^{N}$ and $\{r_i\}_{i=1}^{N}$. Higher alignment indicates that the bundle better distinguishes higher- and lower-reward trajectories. Note that to prevent the generated rubrics from degenerating into simple rules equivalent to outcome rewards,
Equation~\ref{eq:rubric-generation} does not take outcome rewards as input when generating rubric bundles. We finally select the bundle with the highest correlation:
\begin{equation}
m^\star=\arg\max_m\rho^{(m)},
\qquad \mathcal B^\star=\mathcal B^{(m^\star)}.
\label{eq:selected-bundle}
\end{equation}

\subsection{Rubric-Based Advantage Shaping}
\label{sec:token-credit}
The selected rubric bundle serves as privileged context for computing token-level credit. For each policy-generated token $y_{i,\ell}$ at position $\ell$ in trajectory $\tau_i$, let $y_{i,<\ell}$ denote the preceding interaction history, including earlier policy-generated tokens and environment observations. We compare the log-likelihoods assigned by
$\pi_{\mathrm{old}}$ to the same token with and without
the selected rubric:
\begin{align}
\Delta_{i,\ell}
&=\log\pi_{\mathrm{old}}(y_{i,\ell}\mid x,\mathcal B^\star,y_{i,<\ell})
-\log\pi_{\mathrm{old}}(y_{i,\ell}\mid x,y_{i,<\ell}).
\label{eq:likelihood-gap}
\end{align}
Both forward passes evaluate the same pre-sampled trajectories, with no additional sampling. We compute the gap only for policy-generated tokens and mask environment observations. A positive $\Delta_{i,\ell}$ increases the token's advantage, while a negative value decreases it. The resulting token-level advantage is
\begin{equation}
A_{i,\ell}^{\mathrm{token}}=A_i^{\mathrm{GRPO}}+\alpha\Delta_{i,\ell}.
\label{eq:token-adv}
\end{equation}
The coefficient $\alpha\geq 0$ controls the correction magnitude. 
Note that the per-token advantage $A_{i,\ell}^{\mathrm{token}}$ is derived from trajectories generated by the policy itself. For this reason, we categorize ORSD as a self-distillation mechanism that enables fine-grained credit assignment.

\subsection{Policy Optimization}
\label{sec:turn-replay}

We optimize the policy using the token-level advantage
$A_{i,\ell}^{\mathrm{token}}$ defined in Eq.~\eqref{eq:token-adv}.
For each policy-generated token, we define the importance sampling ratio as
\begin{equation}
r_{i,\ell}(\theta)=
\frac{
\pi_\theta(y_{i,\ell}\mid x,y_{i,<\ell})
}{
\pi_{\mathrm{old}}(y_{i,\ell}\mid x,y_{i,<\ell})
}.
\label{eq:policy-ratio}
\end{equation}

The policy is optimized by minimizing the following clipped objective:
\begin{equation}
\begin{split}
\mathcal L_{\mathrm{ORSD}}(\theta)
={}&-\frac{1}{|\mathcal B_{\mathrm{valid}}|}
\sum_{(i,\ell)\in\mathcal B_{\mathrm{valid}}}
\min\Big\{
r_{i,\ell}(\theta)A_{i,\ell}^{\mathrm{token}},
\\
&\qquad
\operatorname{clip}\!\left(
r_{i,\ell}(\theta),1-\varepsilon_c,1+\varepsilon_c
\right)A_{i,\ell}^{\mathrm{token}}
\Big\}
+\eta\mathcal R_{\mathrm{KL}}(\theta).
\end{split}
\label{eq:joint-objective}
\end{equation}

where $\mathcal B_{\mathrm{valid}}$ denotes the set of valid policy-generated tokens in the micro-batch, $\varepsilon_c$ is the clipping threshold, and $\mathcal R_{\mathrm{KL}}$ denotes the KL regularization weighted by $\eta$.

\section{Experiments}
\label{sec:experiments}

\subsection{Experimental Setup}
\label{sec:experimental-setup}

\paragraph{Benchmarks and Metrics.}
We evaluate ORSD on ALFWorld~\citep{shridhar2020alfworld} and WebShop~\citep{yao2022webshop} two long-horizon interactive benchmarks using Qwen3-1.7B and Qwen3-4B~\citep{yang2025qwen3}. For ALFWorld, we report task-wise and overall success rates on the unseen split. For WebShop, we report the official task score and success rate.

\paragraph{Prompting Baselines.}
We include published results for prompting methods that guide interaction without RL training. ReAct~\citep{yao2022react} interleaves reasoning and actions. Reflexion~\citep{shinn2023reflexion} uses reflections on previous attempts to guide later attempts, while ReflAct~\citep{kim2025reflact} relates the current state to the task goal when choosing actions. ReAct w/ Self-Guidance (SG) adds self-generated guidance during interaction~\citep{wang2026selfguide}.

\paragraph{Training Baselines.}

We compare ORSD with several baselines. (a) GRPO: its training data, reward definition, thinking mode, rollout count, and the number of training steps are strictly consistent with ORSD. (b) GiGPO~\citep{feng2025gigpo}: it estimates step-level advantages by grouping actions from similar states. We run GiGPO on both benchmarks with both Qwen3 backbones. (c) Self-distillation methods, including OPSD~\citep{zhao2026opsd}, GRPO+OPSD, and SDAR~\citep{lu2026sdar}. GRPO+OPSD combines GRPO with an auxiliary distillation loss. The sources of published results and their evaluation protocols are specified in the notes to Table~\ref{tab:main-results}.

\paragraph{Training Details.}
We fine-tune all parameters of both backbones with VERL/HybridFlow~\citep{sheng2024hybridflow}. Each run uses four A100 80GB GPUs. Every update contains eight tasks with eight rollouts per task. We train for 300 training steps at a learning rate of $10^{-6}$. WebShop episodes allow for at most 15 assistant actions, while ALFWorld interactions end according to the environment's termination conditions. We generate $M=4$ candidate rubric bundles per task and use $\alpha=0.01$, $\varepsilon_c=0.2$, and $\eta=10^{-3}$ by default. Appendix~\ref{app:configuration} provides the remaining implementation details.

\paragraph{Reward Definition}
\label{sec:reward-design}
For ALFWorld, the training reward is the binary task success indicator:
\begin{equation}
R_{\mathrm{train}}(\tau)=\mathbb{I}[\mathrm{success}],
\label{eq:alfworld-reward}
\end{equation}
These rewards also form the outcome vector used for bundle selection. For WebShop, we use
\begin{align}
R_{\mathrm{train}}(\tau)
&=0.5R_{\mathrm{env}}(\tau)+0.5\mathbb{I}[R_{\mathrm{env}}(\tau)=1]
\nonumber\\
&\quad-0.05\frac{N_{\mathrm{invalid}}(\tau)}{\max(T,1)}
-0.025\frac{N_{\mathrm{loop}}(\tau)}{\max(T,1)} .
\label{eq:webshop-reward}
\end{align}
Here, $R_{\mathrm{env}}(\tau)\in[0,1]$ is the official WebShop task score, and $T$ counts all attempted environment actions, including invalid ones. $N_{\mathrm{invalid}}$ and $N_{\mathrm{loop}}$ count invalid actions and repeated actions that make no progress. This training reward is used to compute GRPO advantages and select rubrics on WebShop. Baselines run in this study use the same reward transformation. Evaluation only uses the original official metrics, namely the task score and success rate.

\subsection{Main Results}
\label{sec:main-results}

\textbf{ORSD achieves the best overall results across both benchmarks and model scales.}
As shown in Table~\ref{tab:main-results}, ORSD ranks first in ALFWorld overall success and in both WebShop metrics with Qwen3-1.7B and Qwen3-4B. On ALFWorld, ORSD exceeds the strongest baseline, GRPO, by 12.12 and 3.36 percentage points in overall success at the two model scales, respectively. On WebShop, ORSD improves success over the strongest baseline by 2.80 percentage points with Qwen3-1.7B, compared with GiGPO, and by 10.90 points with Qwen3-4B, compared with GRPO. The accompanying gains in WebShop score show that the improvement extends to satisfying the requested product attributes.

These results demonstrate the effectiveness of ORSD across household interaction and web shopping tasks. Its overall gains extend beyond GRPO to GiGPO and the reported self-distillation baselines, including SDAR. ORSD derives task-specific process criteria from rollout differences and uses the selected rubric to refine token-level learning signals through paired replay. The improvements over GRPO under the same environment-interaction budget support this approach to extracting additional supervision from existing trajectories. The following ablations examine the contributions of online rubric generation and rubric-reward alignment.

\begin{table}[!t]
\centering
\setlength{\tabcolsep}{1pt}
\definecolor{orsdRow}{HTML}{ECE9FF}
\caption{Performance on ALFWorld and WebShop. Bold indicates the best result for each metric.}
\label{tab:main-results}
\settowidth{\orsdModelColumnOffset}{Qwen3}
\addtolength{\orsdModelColumnOffset}{\tabcolsep}
\sbox{\orsdMainResultsBox}{%
\begin{tabularx}{\textwidth}{@{}c|@{\hspace{5pt}}l*{6}{>{\centering\arraybackslash}X}c*{2}{>{\centering\arraybackslash}X}@{}}
\toprule
& & \multicolumn{7}{c}{ALFWorld} & \multicolumn{2}{c}{WebShop}\\
\cmidrule(lr){3-9}\cmidrule(lr){10-11}
Model & Method & Pick & Look & Clean & Heat & Cool & Pick2 & Overall & Score & Succ.\\
\midrule
\multirow{10}{*}{\shortstack{Qwen3\\1.7B}} & ReAct$^{\S}$ & 24.20 & 12.50 & 0.00 & 0.00 & 0.00 & 12.10 & 10.20 & 41.60 & 4.70\\
 & Reflexion$^{\S}$ & 22.90 & 15.40 & 14.80 & 18.80 & 16.00 & 8.30 & 16.40 & 39.80 & 16.00\\
 & ReflAct$^{\S}$ & 28.60 & 23.10 & 7.40 & 6.30 & 8.00 & 0.00 & 12.90 & 37.50 & 4.40\\
 & ReAct w/ SG$^{\S}$ & 54.20 & 14.30 & 18.50 & 0.00 & 12.10 & 19.00 & 21.10 & 39.70 & 5.40\\
\cmidrule(lr){2-11}
 & OPSD$^{\dagger}$ & 26.30 & 33.30 & 9.10 & 0.00 & 4.50 & 5.30 & 14.10 & 47.40 & 9.30\\
 & GRPO+OPSD$^{\dagger}$ & 38.20 & 50.00 & 30.80 & 28.60 & 30.00 & 21.10 & 32.00 & 70.70 & 38.30\\
 & SDAR$^{\dagger}$ & 73.50 & 25.00 & 76.90 & 33.30 & 40.00 & 36.80 & 53.90 & 76.80 & 58.60\\
 & GRPO & 68.75 & \textbf{72.22} & 50.81 & 54.35 & 36.90 & \textbf{50.00} & 53.92 & 72.20 & 55.40\\
 & GiGPO & 79.20 & 27.80 & \textbf{77.40} & 34.80 & 38.10 & 41.20 & 53.00 & 76.80 & 60.20\\
\rowcolor{orsdRow}
\cellcolor{white} & \textbf{ORSD (ours)} & \textbf{81.25} & 68.06 & 69.35 & \textbf{79.35} & \textbf{53.57} & 41.18 & \textbf{66.04} & \textbf{78.60} & \textbf{63.00}\\
\midrule
\multirow{7}{*}{\shortstack{Qwen3\\4B}} & ReAct$^{\S}$ & 42.40 & 16.70 & 0.00 & 8.30 & 9.10 & 28.00 & 20.30 & 31.10 & 11.30\\
 & Reflexion$^{\S}$ & 68.60 & 38.50 & 37.00 & 25.00 & 32.00 & 4.20 & 37.10 & 57.80 & 24.00\\
 & ReflAct$^{\S}$ & 74.30 & 38.50 & 25.90 & 25.00 & 32.00 & 16.70 & 38.60 & 35.10 & 16.00\\
 & ReAct w/ SG$^{\S}$ & 87.80 & 50.00 & 29.20 & 58.30 & 27.30 & \textbf{80.00} & 58.60 & 42.70 & 17.60\\
\cmidrule(lr){2-11}
 & GRPO & 83.33 & \textbf{86.11} & 73.39 & \textbf{72.83} & 72.62 & 63.24 & 74.07 & 82.69 & 66.15\\
 & GiGPO & \textbf{88.20} & 50.00 & \textbf{96.20} & 52.40 & 65.00 & 57.90 & 73.40 & 81.80 &65.60\\
\rowcolor{orsdRow}
\cellcolor{white} & \textbf{ORSD (ours)} & 83.33 & 77.78 & 87.10 & 70.65 & \textbf{83.33} & 63.24 & \textbf{77.43} & \textbf{89.02} & \textbf{77.05}\\
\bottomrule
\end{tabularx}%
}%
\usebox{\orsdMainResultsBox}%
\llap{\hbox to \wd\orsdMainResultsBox{%
  \hspace*{\orsdModelColumnOffset}%
  \vrule width \arrayrulewidth height \ht\orsdMainResultsBox depth \dp\orsdMainResultsBox
  \hfill}}
\par\smallskip
\begin{minipage}{\textwidth}
$\S$: Prompting results from \emph{Co-Evolution of Policy and Internal Reward for Language Agents}~\citep{wang2026selfguide}. $\dagger$: OPSD, GRPO+OPSD, and SDAR results from \emph{Self-Distilled Agentic Reinforcement Learning}~\citep{lu2026sdar}. Published results retain their source evaluation protocols.
\end{minipage}
\end{table}

\subsection{Ablation Study}
\label{sec:ablations}
In this section, we use ablation experiments to examine the effectiveness of ORSD's components: online rubric generation, the number of candidate bundles and their selection, and the advantage shaping coefficient $\alpha$. All ablations use WebShop. We study rubric generation and selection with both Qwen3 backbones and vary $\alpha$ with Qwen3-1.7B.

\paragraph{Effectiveness of Online Rubric Generation.}
To evaluate the effectiveness of online rubric generation, we construct an \textit{Offline Rubric} variant. Before RL training, the initial policy generates one task-specific rubric bundle for each task, and these rubrics remain fixed throughout training. As shown in Table~\ref{tab:ablations}, Offline Rubric consistently improves over GRPO on both model scales, confirming that task-specific rubric supervision is already beneficial even when generated only once. However, it still falls short of online ORSD by 4.80 percentage points in success rate for both backbones. This gap highlights the value of refreshing the rubric from the current policy's rollouts, so that the supervision remains aligned with the policy's changing behavior and failure modes during training. Importantly, both variants generate rubrics automatically and require no manually authored criteria.

\paragraph{Effectiveness of Rubric-Reward Alignment.}
ORSD selects candidate rubric bundles based on their alignment with outcome rewards. We examine this design through two ablations. \textit{ORSD w/o Rubric-Reward Alignment} generates a single bundle and uses it directly, omitting verifier scoring and candidate selection. \textit{Random Rubric} retains the same candidate count as full ORSD ($M=4$) but replaces alignment-based selection with random selection. Table~\ref{tab:ablations} shows that full ORSD outperforms both variants on both model scales. Relative to the single-bundle variant, ORSD improves success by 2.15 and 9.45 percentage points on Qwen3-1.7B and Qwen3-4B, respectively. With the candidate count held fixed, alignment-based selection improves success over random selection by 3.40 and 7.15 percentage points, with corresponding score gains of 5.36 and 4.08 points. These consistent gains demonstrate the effectiveness of rubric-reward alignment in selecting process supervision for policy training, with the selected rubrics serving as privileged context for token-level advantage shaping.

\begin{table}[!htbp]
\centering
\caption{Component ablations on WebShop with two policy backbones. Baseline denotes GRPO. ORSD w/o Rubric--Reward Alignment generates one bundle and omits verifier scoring and candidate selection. Rubric-based variants use $\alpha=0.01$.}
\label{tab:ablations}
\begin{tabular}{@{}lcccc@{}}
\toprule
& \multicolumn{2}{c}{Qwen3-1.7B} & \multicolumn{2}{c}{Qwen3-4B}\\
\cmidrule(lr){2-3}\cmidrule(lr){4-5}
Method & Score & Success (\%) & Score & Success (\%)\\
\midrule
Baseline & 72.20 & 55.40 & 82.69 & 66.15\\
Offline Rubric & 74.35 & 58.20 & 85.39 & 72.25\\
ORSD w/o Rubric--Reward Alignment & 77.53 & 60.85 & 83.86 & 67.60\\
Random Rubric & 73.24 & 59.60 & 84.94 & 69.90\\
\textbf{ORSD ($M=4$)} & \textbf{78.60} & \textbf{63.00} & \textbf{89.02} & \textbf{77.05}\\
\bottomrule
\end{tabular}
\end{table}

\begin{figure}[!ht]
\centering
\includegraphics[width=0.72\textwidth]{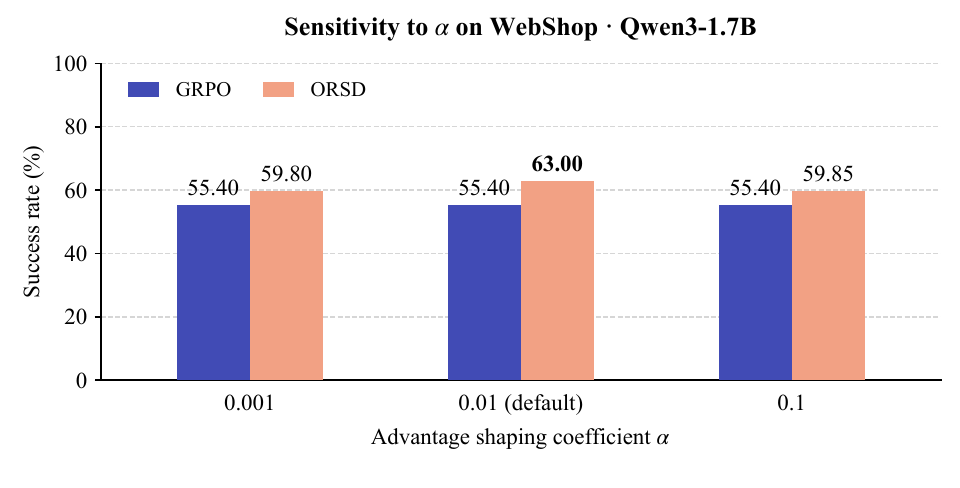}
\caption{Sensitivity to $\alpha$ on WebShop with Qwen3-1.7B. The default is $\alpha=0.01$. Each setting uses the same GRPO baseline (55.40\% success).}
\label{fig:alpha-sensitivity}
\end{figure}

\paragraph{Sensitivity to $\alpha$.}
Figure~\ref{fig:alpha-sensitivity} compares the default ORSD coefficient $\alpha=0.01$ with $\alpha=0.001$ and $\alpha=0.1$ on WebShop using Qwen3-1.7B. We report the success rate and use the same GRPO result as the baseline at each setting; GRPO corresponds to removing the correction in Eq.~(\ref{eq:token-adv}). All three coefficients outperform GRPO, with the default $\alpha=0.01$ achieving the highest success rate of 63.00\%. These results indicate that $\alpha=0.01$ provides the most effective correction strength among the tested values, supporting its use as the default setting.

\subsection{Further Analysis}
\label{sec:rubric-examples}
\label{sec:mechanistic-diagnostics}

\paragraph{Training Dynamics}

\begin{figure}[!htbp]
\centering
\includegraphics[width=\textwidth]{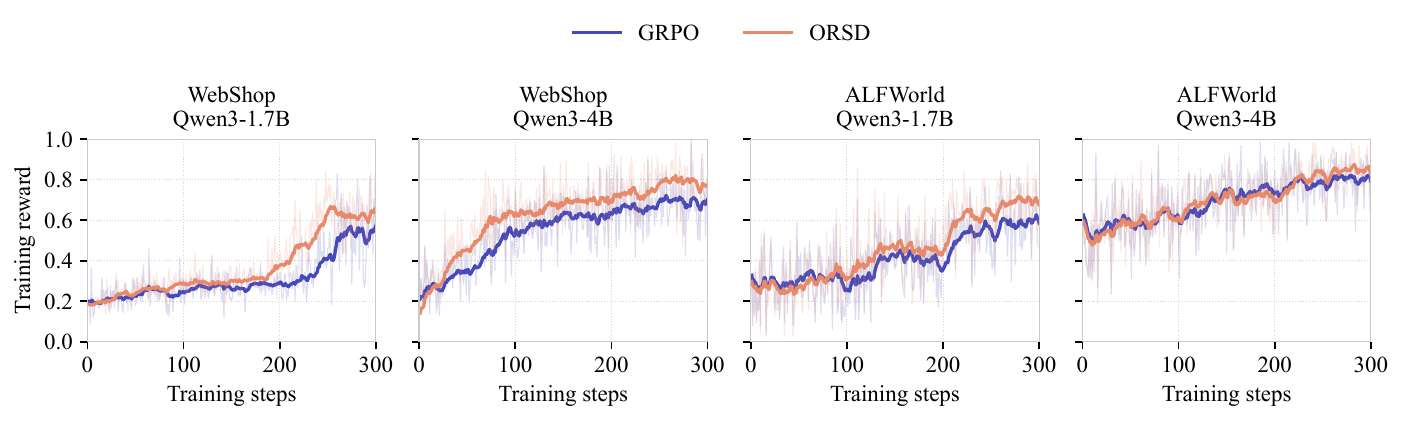}
\caption{Training reward curves comparing ORSD with GRPO on WebShop and ALFWorld using Qwen3-1.7B and Qwen3-4B. Both methods use the same environment-interaction budget.}
\label{fig:learning-curves}
\end{figure}

Figure~\ref{fig:learning-curves} compares the training rewards of ORSD and GRPO on both benchmarks and model scales. On WebShop, ORSD improves earlier with Qwen3-1.7B and maintains higher rewards through most of training with Qwen3-4B. On ALFWorld, the advantage is more pronounced at 1.7B and emerges later at 4B. Across all four settings, ORSD achieves higher mean rewards over the final 300 optimization steps under the same environment-interaction budget. Together with the gains in Table~\ref{tab:main-results}, these results support more effective use of sampled trajectories: ORSD reuses rollout experience to derive fine-grained supervision through self-distillation, improving learning without additional environment rollouts.

\paragraph{Analysis of Generated Rubric Bundles}
Table~\ref{tab:rubric-examples} compares a selected WebShop rubric with a rejected candidate. The selected bundle specifies product attributes and a concrete price constraint, providing more informative task-specific privileged context than the rejected candidate's general criteria. This comparison illustrates how rubric-reward alignment selects more informative privileged context to guide token-level advantage shaping.

\begin{table*}[!t]
\centering
\caption{WebShop rubric comparison. The left column contains a rubric saved at step 300; the right column contains a rejected proposal.}
\label{tab:rubric-examples}
\begin{tabularx}{\textwidth}{@{}>{\raggedright\arraybackslash}X>{\raggedright\arraybackslash}X@{}}
\toprule
Saved rubric & Rejected rubric\\
\midrule
\multicolumn{2}{@{}p{\textwidth}@{}}{WebShop: Buy a women's top with the requested attributes, color, and size for less than \$30.}\\
\midrule
1. Identify the correct product category and specific product attributes (e.g., slim fit, loose fit, color, size, price) needed to find the desired item. & 1. Identify the appropriate product category and the key attributes specified in the user instruction.\\
2. Ensure the search query includes all relevant keywords and filters to narrow down the results effectively. & 2. Use relevant product keywords and available search or filtering options to narrow down candidate items.\\
3. Search using the relevant product attributes and narrow the candidates accordingly. Do not commit to a product based only on a partial match in the title or description. & 3. Select the appropriate product variant, such as the requested size, color, or other configurable attributes.\\
4. Verify that the selected product meets the price criteria (lower than \$30.00) before proceeding to purchase. & 4. Verify that the selected product satisfies the important constraints in the instruction, including product attributes and price requirements.\\
5. Confirm that the product is correctly selected and the purchase action is initiated. & 5. Confirm that the final product matches the user's request and complete the purchase successfully.\\
\bottomrule
\end{tabularx}
\end{table*}

\section{Discussion and Limitations}
\label{sec:discussion}
 
\paragraph{Rubric Knowledge is not explicitly accumulated across updates.}
ORSD generates task-specific rubrics from the current rollout
group at each update, but does not maintain a persistent
repository of previously discovered criteria. How to retain,
refine, and reuse useful rubrics across training stages and
related tasks remains an open question. Extending ORSD with
a rubric memory could support cumulative process knowledge
while preserving adaptation to the evolving policy.

\paragraph{Outcome alignment does not fully resolve credit assignment.}
ORSD selects rubric bundles by their correlation with
trajectory-level rewards. This objective favors rubrics
that distinguish stronger trajectories, but does not
directly optimize their usefulness for assigning credit
within a trajectory. Future work could develop selection
objectives that account for how rubric guidance affects
individual decisions and subsequent policy improvement.

\section{Conclusion}

We introduced \method{} (ORSD) to improve credit assignment in LLM agent reinforcement learning. ORSD generates task-specific process rubrics online from differences among on-policy trajectories, without manually authored or fixed offline criteria. Rubric-reward alignment selects the bundle whose satisfaction scores best correlate with outcomes. Through privileged paired replay, the selected rubric converts token-level likelihood gaps into token-level advantage corrections, providing fine-grained supervision without additional environment interactions. Experiments on ALFWorld and WebShop with Qwen3-1.7B and Qwen3-4B show the best overall performance among the compared methods, while ablations validate online rubric generation and candidate selection. These results highlight the value of deriving adaptive process supervision from the policy's own experience for credit assignment in long-horizon agent tasks.

\clearpage

\appendix
\section{Algorithm}
\label{app:algorithm}

\begin{enumerate}
    \item Let $\pi_{\mathrm{old}}$ denote the policy used for trajectory sampling and paired replay in the current update.
    \item For every task $x$, sample a group of $N=8$ trajectories $\mathcal T$ and compute $A_{i}^{\mathrm{GRPO}}$.
    \item Aggregate the task and all trajectories in the group into one rubric-generation prompt, then independently sample four complete rubric bundles with 5-8 criteria each using the current policy.
    \item Use the frozen verifier to score every criterion in each complete bundle. Average all criterion scores within each trajectory, then compute bundle-level Spearman correlation with group outcomes and select the complete bundle with the highest score.
    \item Replay each stored action turn with and without $\mathcal B^\star$. Check response masks and token alignment, compute the raw $\Delta_{i,\ell}$, and add the valid-token correction in Eq.~(\ref{eq:token-adv}).
    \item Accumulate the policy loss and reference-policy KL regularizer in Eq.~(\ref{eq:joint-objective}); update the shared policy once.
\end{enumerate}

\section{Prompt Templates}
\label{app:prompt}

One aggregated prompt supplies the current policy with the task and all action-observation transcripts in the rollout group. With the group-hint instruction, the prompt requests a complete process rubric containing 5-8 observable criteria. The policy samples four independent complete proposals from the same prompt, and each proposal remains intact throughout selection.

For verification, the frozen verifier receives the task, the rollout transcripts, and one candidate bundle. Each criterion is scored independently from visible evidence and receives a satisfaction score in $[0,1]$. All criterion scores in the bundle are averaged within each trajectory, then correlated with group outcomes across trajectories using Spearman correlation.

The following templates specify rubric generation and policy interaction in the two environments. Blue placeholders denote fields populated for each task. For rubric generation, $k=8$ and the previous-attempts field contains the same-task rollout group.

\begin{orsdprompt}{orsdPromptBlue}{Rubric Generation Prompt}
\textbf{[SYSTEM]}

You are a task coach who writes process rubrics for an interactive environment.

You will see one task and several previous attempts at the same task.\newline Use them to write useful process criteria for a NEW agent that has never seen those attempts.

Write 5 to \promptvar{k} short rubric criteria.

\textbf{Useful rubric criteria can describe:}\newline - important subgoals needed to complete the task,\newline - useful ordering of subgoals,\newline - mistakes or no-progress behavior to avoid,\newline - what must be true before the task is finished.

\textbf{Rules:}\newline - Use only information supported by the task or the attempts.\newline - Do not mention attempts, trajectories, rollouts, rewards, sources, evidence, or IDs.\newline - Do not copy a complete action sequence.\newline - Do not invent object locations or hidden state.\newline - Keep each rubric criterion short, concrete, and useful by itself.\newline - Do not explain your reasoning.

\textbf{Output only a numbered process rubric:}\newline 1. ...\newline 2. ...\newline 3. ...\newline 4. ...\newline 5. ...

\textbf{[USER]}

\textbf{Task:}\newline
\promptvar{problem}

\textbf{Previous attempts:}\newline
\promptvar{attempts}

Write a useful process rubric for a new agent solving this same task.
\end{orsdprompt}

\begin{orsdprompt}{orsdPromptPurple}{ALFWorld Interaction Prompt}
\textbf{[SYSTEM]}

You are acting in ALFWorld. Observe the environment and its valid actions, maintain task state, and reply with exactly `ACTION: \textless{}one currently valid action\textgreater{}`.

\textbf{[USER]}

Goal: \promptvar{goal}
\end{orsdprompt}

\begin{orsdprompt}{orsdPromptGreen}{WebShop Interaction Prompt}
\textbf{[SYSTEM]}

You are an agent interacting with a virtual text-based web shopping environment to test out your ability. Your job is to follow the Instruction provided and mimic the steps to buy the item that are closest to the Instruction provided.

\textbf{Action Format:}\newline You should give both the action\_name and action\_arg like the format `action\_name[action\_arg]`. You can execute two types of actions, search and click.\newline - When the button `[button] Search [button\_]` is available in the current observation, you can execute the action `search[xxx]` (you should type the query you want to search in the square brackets here).\newline - You can click buttons `[button] xxx [button\_]` that is available in the current observation, by execute the action `click[xxx]`.

Below are some examples of action formats.\newline - ACTION: search[white shoes]\newline - ACTION: click[Buy Now]

\textbf{Notes:}\newline At each step, you should first think then perform action to fulfill the instruction. After thinking, output the action on exactly one line in the format `ACTION: action\_name[action\_arg]`.\newline You should ALWAYS take one action each step.\newline You should finish the task and buy the item within 15 steps.\newline DO NOT try to interact with the user at anytime. Finish the task and buy the item by yourself.

\textbf{[USER]}

\promptvar{instruction}
\end{orsdprompt}

\section{Implementation Details}
\label{app:configuration}

\begin{table}[!htbp]
\caption{Optimization and sampling configuration.}
\centering
\begin{tabularx}{\textwidth}{@{}>{\raggedright\arraybackslash}p{0.38\textwidth}>{\raggedright\arraybackslash}X@{}}
\toprule
Parameter & Value \\
\midrule
Base models & Qwen3-1.7B; Qwen3-4B \\
Optimization & Full-parameter GRPO \\
Hardware & 4 $\times$ A100 80GB \\
Prompts / rollouts per prompt & 8 / 8 \\
Optimizer steps & 300 \\
Rollout group size & 8 (ALFWorld and WebShop) \\
Learning rate & $10^{-6}$ \\
Clipping parameter $\varepsilon_c$ & $0.2$ \\
KL regularizer / coefficient & low-variance estimator / $10^{-3}$ \\
Token-credit coefficient $\alpha$ & $0.01$ \\
Rubric candidates per task & 4 complete bundles \\
Criteria per bundle & 5-8 \\
Bundle selection & highest Spearman correlation with group outcomes \\
Policy micro-batch size per GPU & ALFWorld: 2; WebShop: 1 \\
Replay eligibility & selected bundle; valid alignment and finite scores \\
Frozen verifier & Qwen3.5-Plus via DashScope API \\
Verifier decoding & temperature $0$, thinking disabled, JSON only; 1,536-token cap \\
Policy / privileged-replay thinking & on \\
Generator thinking / prompt & off / aggregated numbered rubric \\
Generator temperature / top-$p$ / cap & $0.3$ / $0.9$ / 4,096 \\
Verifier concurrency / retries / cap & 8 / 4 / 1,536 \\
Validation sampling & 4 trajectories per task, temperature $0.6$ \\
Framework & VERL 0.8, FSDP, async vLLM rollout \\
\bottomrule
\end{tabularx}
\end{table}

\begin{table}[!htbp]
\caption{Environment-specific configuration. Full history denotes the complete sequence of previous reasoning responses, actions, and observations.}
\label{tab:environment-config}
\centering
\begin{tabular}{lcc}
\toprule
Parameter & ALFWorld & WebShop \\
\midrule
Evaluation outcome & official \texttt{won} & official score and score $=1$ \\
Episode horizon & environment termination & 15 assistant actions \\
Response budget & 16,384 & 30,720 \\
Interaction history & full history & full history \\
Bundle-selection outcome & official success & behavior-regularized return \\
Training reward & Eq.~(\ref{eq:alfworld-reward}) & Eq.~(\ref{eq:webshop-reward}) \\
Invalid-action coefficient & $0$ & $0.05$ \\
No-progress coefficient & $0$ & $0.025$ \\
vLLM memory utilization & $0.50$ & $0.72$ \\
Agent-loop workers & 8 & 4 \\
\bottomrule
\end{tabular}
\end{table}

\paragraph{Evaluation and learning curves.}
We evaluate every 50 optimizer steps on fixed held-out tasks. Comparison runs in this study use the same evaluation checkpoints and training seeds. Figure~\ref{fig:learning-curves} plots training rewards against training steps. Each update collects 64 environment trajectories, so step 300 corresponds to 19,200 training trajectories. Verifier calls, replay computation, and wall-clock time are recorded separately.

\subsection{Policy Loss Reduction}
\label{app:loss-reduction}
The main text uses $\mathcal A_{\mathrm{mb}}$ to denote the masked micro-batch reduction. Let $D$ be the number of data-parallel workers, $G$ the number of gradient accumulation steps per worker, and $\mathcal C_{d,g}$ the trajectories in the corresponding micro-batch, indexed across tasks for this reduction. For any token-level quantity $f$,
\begin{equation}
\mathcal A_{\mathrm{mb}}[f]
=\frac{1}{DG}\sum_{d=1}^{D}\sum_{g=1}^{G}
\frac{\sum_{i\in\mathcal C_{d,g}}\sum_{\ell}
b_{i,\ell}f_{i,\ell}}
{\sum_{i\in\mathcal C_{d,g}}\sum_{\ell}b_{i,\ell}}.
\label{eq:microbatch-reduction}
\end{equation}
Here, $b_{i,\ell}$ is a binary mask for valid response tokens, excluding prompts, environment observations, and padding. This reduction first averages over tokens within each micro-batch, then over micro-batches and workers; it does not separately average over tasks or turns. The term $\varepsilon_{\mathrm{num}}$ ensures numerical stability when computing GRPO advantages.

\subsection{Paired Replay Data}

Paired replay uses records saved at each generation turn. A record contains the turn index, context messages and token IDs, complete response token IDs, action-only response history, subsequent observation, response mask, and rollout log probabilities. Plain replay packs $c_t+y_t$ independently for each turn. Rubric-conditioned replay inserts the selected complete bundle into $c_t$ and reuses the same target response. The response mask selects the log probabilities used by the correction. The implementation mask $e_{i,\ell}$ equals one when the tokenization round-trip succeeds, response tokens align, and both log-likelihoods are finite. A token with $e_{i,\ell}=0$ retains its GRPO advantage. The main-text equations omit this implementation mask. The policy ratio continues to use the original context and saved rollout log probabilities.

\subsection{Training entry points and run records}
\label{app:unresolved}

Two launchers under \texttt{examples/agent\_rubric2} define the WebShop group-hint and ALFWorld first-version group-hint configurations. Both aggregate all trajectories in a same-task rollout group, sample four complete bundles with the current policy, select one complete proposal, and apply the raw additive token correction. GRPO uses the same four-GPU, eight-task, eight-rollout, 300-step protocol with rubric generation and token correction disabled.

Rubric generation uses temperature $0.3$, top-$p=0.9$, and a 4,096-token cap. Verification calls \texttt{qwen3.5-plus} with concurrency 8, at most four retries, and a 1,536-token output cap. Each run record contains the API identifier and date, resolved configuration, task IDs, seeds, saved trajectories, bundle scores, and checkpoint evaluations. Runs are grouped under \texttt{verl\_rubric\_3}.

Both replay branches preserve the per-turn context and response IDs, masks, chat templates, temperature settings, and rubric insertion rules. Before applying the advantage correction, token scores are written back to their original response positions. The micro-batch loss reduction and reference-policy KL settings match GRPO.

\section{Mathematical Form of the Token Correction}
\label{app:math}

\paragraph{Effect on trajectory credit.}
For $n_{i}=\sum_{\ell}b_{i,\ell}>0$, the actual mean advantage is
\[
\frac{1}{n_{i}}\sum_{\ell}b_{i,\ell}A^{\mathrm{token}}_{i,\ell}
=A_{i}^{\mathrm{GRPO}}+
\frac{\alpha}{n_{i}}\sum_{\ell}b_{i,\ell}e_{i,\ell}\Delta_{i,\ell}.
\]
The raw correction can therefore change both the trajectory mean and individual token signs. Its scale is controlled by $\alpha$ and the replay-alignment mask; no additional normalization or redistribution is applied.

\paragraph{Self-distillation through advantage shaping.}
Both terms in $\Delta_{i,\ell}$ are computed by the same $\pi_{\mathrm{old}}$. ORSD applies stop-gradient, denoted by $\sg$ below, to the likelihood gap and includes it only in $A_{i,\ell}^{\mathrm{token}}$. At the start of an update, the additional contribution to the policy-loss gradient is
\begin{equation}
-\alpha\,\mathcal A_{\mathrm{mb}}\!\left[
e_{i,\ell}\,\sg[\Delta_{i,\ell}]\,
\nabla_\theta\log\pi_\theta(y_{i,\ell}\mid x,y_{i,<\ell})
\right]_{\theta=\theta_{\mathrm{old}}}.
\end{equation}
Here, $\mathcal A_{\mathrm{mb}}$ denotes the masked micro-batch reduction in Eq.~(\ref{eq:microbatch-reduction}), applied to the enclosed token-level quantity.
Privileged context changes the weight assigned to each sampled token in the RL update. ORSD implements privileged self-distillation through advantage shaping; $\mathcal L_{\mathrm{total}}$ contains no auxiliary KL or cross-entropy matching objective. The likelihood gap measures the policy's response to the selected process context and does not establish token correctness.

\end{document}